\documentclass[sigconf]{acmart}

\usepackage{mathtools}
\usepackage{subfigure}

\usepackage{amsmath,array}
\usepackage{amsthm}
\usepackage{comment}
\usepackage{mathtools}
\usepackage{datetime}
\usepackage{graphicx}
\usepackage{cuted}
\usepackage{tikz,pgfplots,pgfplotstable}
\usepackage{xfrac}
\usepackage{mwe}
\usetikzlibrary{patterns}
\usepackage{verbatim}
\usepackage{algorithm,algorithmic}
\usepackage{bm}
\usepgfplotslibrary{patchplots}

\usetikzlibrary{patterns}

\usepackage{color}

\newcommand\tableSize{0.67}

\AtBeginDocument{%
  \providecommand\BibTeX{{%
    \normalfont B\kern-0.5em{\scshape i\kern-0.25em b}\kern-0.8em\TeX}}}

\copyrightyear{2022}
\acmYear{2022}
\setcopyright{rightsretained}

\acmConference[WiseML '22]{Proceedings of the 2022 ACM Workshop on Wireless Security and Machine Learning}{May 19, 2022}{San Antonio, TX, USA.}
\acmBooktitle{Proceedings of the 2022 ACM Workshop on Wireless Security and Machine Learning (WiseML '22), May 19, 2022, San Antonio, TX, USA}
\acmDOI{10.1145/3522783.3529524}
\acmISBN{978-1-4503-9277-8/22/05}

\usepackage{etoolbox}
\makeatletter
\patchcmd{\maketitle}{\@copyrightpermission}{
   \begin{minipage}{0.4\columnwidth}
     \href{https://creativecommons.org/licenses/by-nc/4.0/}{\includegraphics[width=0.95\textwidth]{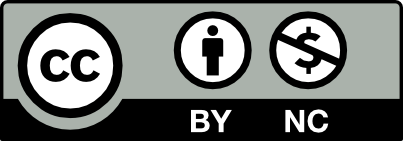}}
   \end{minipage}\hfill
   \begin{minipage}{0.6\columnwidth}
     \href{https://creativecommons.org/licenses/by-nc/4.0/}{This work is licensed under a Creative Commons Attribution-NonCommercial International 4.0 License.}
   \end{minipage}

   \vspace{5pt}
}{}{}

\makeatother

\begin{document}
\fancyhead{}

\title{Automatic Machine Learning for Multi-Receiver CNN Technology Classifiers}

\author{Amir-Hossein Yazdani-Abyaneh}
\email{yazdaniabyaneh@email.arizona.edu}
\affiliation{%
  \institution{ECE Department, University of Arizona}
  \streetaddress{P.O. Box 1212}
  \city{Tucson}
  \state{Arizona}
  \country{USA}
}

\author{Marwan Krunz}
\email{krunz@email.arizona.edu}
\affiliation{%
  \institution{ECE Department, University of Arizona}
  \streetaddress{P.O. Box 1212}
  \city{Tucson}
  \state{Arizona}
  \country{USA}
}

\begin{abstract}

 \textit{Convolutional Neural Networks (CNNs)} are one of the most studied family of deep learning models for signal classification, including modulation, technology, detection, and identification. In this work, we focus on technology classification based on raw I/Q samples collected from multiple synchronized receivers. As an example use case, we study protocol identification of  Wi-Fi, LTE-LAA, and 5G NR-U technologies that coexist over the 5 GHz \textit{Unlicensed National Information Infrastructure (U-NII)} bands. Designing and training accurate CNN classifiers involve significant time and effort that  goes to fine-tuning a model's architectural settings (e.g., number of convolutional layers and their filter size) and determining the appropriate hyperparameter configurations, such as learning rate and batch size. We tackle the former by defining architectural settings themselves as hyperparameters. We attempt to automatically optimize these architectural parameters, along with other preprocessing (e.g., number of I/Q samples within each classifier input) and learning hyperparameters, by forming a \textit{Hyperparameter Optimization (HyperOpt)} problem, which we solve in a near-optimal fashion using the \textit{Hyperband} algorithm. The resulting near-optimal CNN (OCNN) classifier is then used to study classification accuracy for OTA as well as simulations datasets, considering various SNR values. We show that using a larger number of receivers to construct multi-channel inputs for CNNs does not necessarily improve classification accuracy. Instead, this number should be defined as a preprocessing hyperparameter to be optimized via Hyperband.  OTA results reveal that our OCNN classifiers improve classification accuracy by $24.58\%$  compared to manually tuned CNNs. We also study the effect of min-max normalization of I/Q samples within each classifier's input on generalization accuracy over simulated datasets with SNRs other than training set's SNR, and show an average of $108.05\%$ improvement when I/Q samples are normalized.
\end{abstract}



\maketitle

\section{Introduction}
Wi-Fi, LTE Licensed Assisted Access (LAA) and 5G New Radio Unlicensed (NR-U) can operate over the 5 GHz Unlicensed National Information Infrastructure (U-NII) bands. To coexist harmoniously, these technologies use variants of Carrier Sense Multiple Access with Collision Avoidance (CSMA/CA) to access unlicensed channels. However, the standards for each technology specify different CSMA/CA parameters, e.g., maximum backoff value, clear channel access (CCA) thresholds, etc., resulting in heterogeneous channel access behavior \cite{chen2016coexistence,muhammad20205g}. Studies have shown that for these  technologies to achieve fair coexistence, they have to adapt their channel access parameters and schedule their transmissions based on the transmission activities of all coexisting technologies \cite{gao2020achieving}. Fair coexistence can be greatly facilitated via the classification of active transmissions into their respective technologies, which require technology-specific receivers that search for distinct features within the received waveforms  (e.g., 802.11 preamble, LTE OFDM grid structure, etc.). However, to ease the burden of having multiple technologies implemented and interconnected on commercial devices, an alternative approach adopted in this paper, is to use a universal classifier based on Deep Neural Networks (DNNs), such as \textit{Convolutional Neural Networks (CNNs)}.\\
CNN-based signal classification has been studied to fulfill  different goals, including modulation classification \cite{o2016convolutional,osheaJSAC18}, technology (protocol) identification \cite{amirinfo, bitar2017wireless}, transmitter authentication \cite{jian2020IoT, sankhe2019no}, LTE spectrogram generation \cite{roy2019generative}, and generation of synthetic modulated waveforms via Generative Adversarial Networks (GANs) \cite{patel2020data}. These classifiers suggest classes based on received raw I/Q samples. Directly working with I/Q samples bypasses the need for specialized signal processing (e.g., correcting for carrier frequency offset and channel effects), enabling classification to be done within microseconds \cite{soltani2019real}.\\
In this paper, we focus on unlicensed transmissions in the 5 GHz U-NII frequency bands, where Wi-Fi, LTE-LAA, and 5G NR-U technologies coexist.  Gaining insight from prior research conducted on signal classification using deep learning models, we also choose CNNs as our base classifiers and the baseband I/Q samples as inputs for such classifiers. In a rich scattering environment, including the 5 GHz U-NII bands, multi-path fading results in sufficient spatial diversity between the I/Q samples collected at differently located receivers, including multiple receive chains of the same device. Accordingly, we consider a classifier whose I/Q samples are obtained from multiple synchronized receivers.  This is analogous to an image classification problem, where images are fed to CNNs through multiple color channels, i.e., red, green, and blue \cite{bishop2006pattern}. We study the implications of employing multiple I/Q streams on the classification accuracy.\\
Much time and effort go into  manually designing the architecture of a CNN classifier and tuning its hyperparameters (e.g., learning rate, batch size, etc.). Using \emph{Automatic Machine Learning (AutoML)} techniques, specifically \emph{Hyperparameter Optimization (HyperOpt)} \cite{hutter2019automated}, we define the architectural settings of our CNN model as hyperparameters, and along with other learning and preprocessing hyperparameters, run a HyperOpt algorithm called \emph{Hyperband} \cite{li2017hyperband} to find near-optimal hyperparameters and architectural design settings. Hyperband has shown a $20 \times$  increase in performance (e.g., validation loss) over random search, which is often used for HyperOpt problems with high-dimensional spaces (e.g., DNN design). Random search is often used instead of exhaustive search, which has an exponential complexity.  \\
Considering different ranges of SNR values, we generate two datasets, one obtained over the air (OTA) via USRP experiments and the other using MATLAB simulations. OTA results show that our near-optimal CNNs improve classification accuracy by $24.58\%$, on average, compared to manually tuned CNN classifiers. For our simulation datasets, we show that normalizing the classifier's inputs improves generalization accuracy over  datasets with  SNR values other than the training set's SNR by $108.05\%$. Moreover, our evaluation results suggest that using the maximum available number of receivers to collect I/Q samples from does not improve classification accuracy for some CNN architectures, and this number should be set as a hyperparameter to be optimized within HyperOpt.\\
The paper is organized as follows. Section \ref{sc:related_works} provides a summary of related works. In Section \ref{sc:multi_sc}, we introduce our multi-receiver signal classification problem.
 HyperOpt, Hyperband, and hyperparameter settings of near-optimal CNN structures are presented in Section \ref{sc:automl}. We provide our evaluation results in Section \ref{sc:evaluations} and conclude the paper in Section \ref{sc:conclusion}.
\section{Related Works}\label{sc:related_works}
Zhang et al. \cite{amirinfo} studied a similar technology classification problem of Wi-Fi, LTE, and 5G signals, where they evaluated the classification performance of CNNs and LSTMs using I/Q samples received from a single receiver. In \cite{bitar2017wireless}, Bitar et al. studied the classification of I/Q samples for Bluetooth, ZigBee, and 802.11n in the context of IoT.
 O'Shea et al. \cite{o2016convolutional} studied the problem of modulation classification  and showed improved classification performance for CNNs compared with conventional ML classification algorithms (e.g., SVMs and decision trees). Their work was extended in \cite{osheaJSAC18} by studying different wireless communication imperfections, such as carrier frequency offset (CFO) and symbol rate offset (SRO) in SDR experimental setups.  Shi et al. \cite{shi2019deep} considered primary/secondary user (PU/SU) signal classification in a dynamic spectrum access scenario as a form of modulation classification problem, where PUs and SUs adopt different modulations for their transmissions, and utilized CNN classifiers. Sankhe et al. \cite{sankhe2019no} proposed a transmitter authentication framework for identifying  a set of known transmitters using the fact that I/Q samples transmitted by different transmitters are  affected differently by hardware impairments (e.g., CFO). The authors in \cite{wang2020deep} used multiple streams of I/Q samples from different receivers for modulation classification. They trained individual CNNs for individual receivers, and used cooperative decision rules (e.g., ensemble average) to suggest a final classification outcome. In contrast, in our work, we form a single  input from all I/Q streams (i.e., multi-channel inputs) and feed it to one CNN classifier.  All the above works used manually tuned architectural and hyperparameter settings.\\
\section{Multi-Receiver Signal Classification}\label{sc:multi_sc}
\begin{figure}[h]
 \centering
  \includegraphics[scale=0.4]{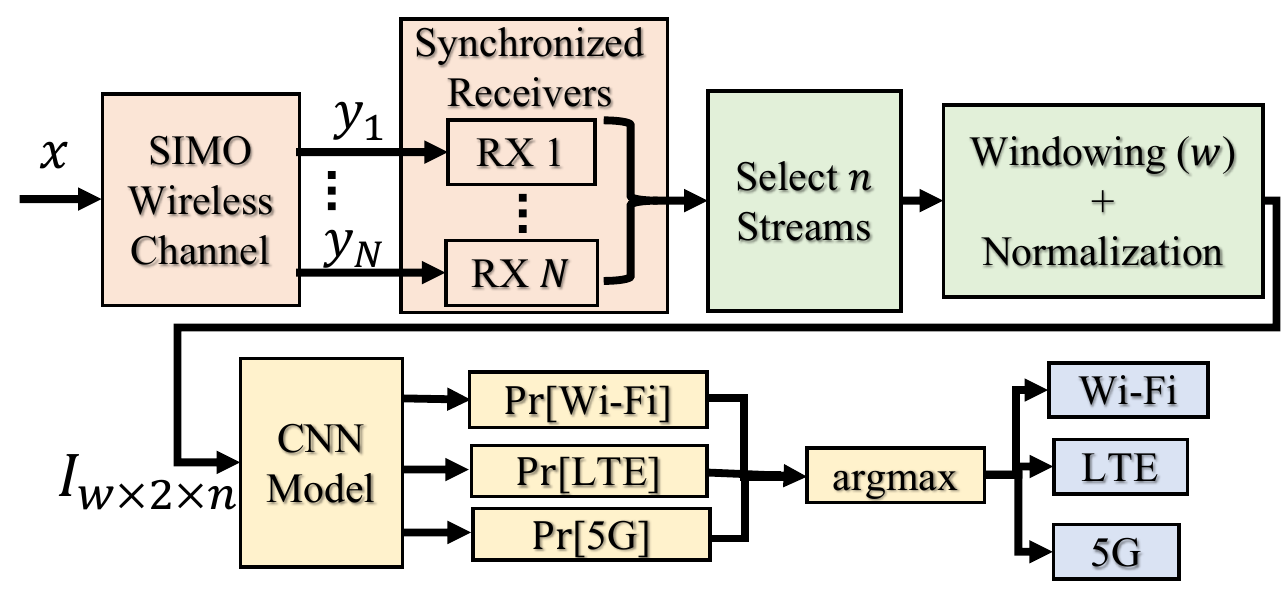}
  \caption{Classification pipeline.}
  \label{fig:pipeline}
\end{figure}

We consider a technology classification problem, where several Wi-Fi, LTE-LAA, and 5G NR-U transmitters coexist over a shared channel.  We assume an interleaving \textit{spectrum sharing} scenario, where at any given time, only one type of transmission is possible. At the classifier, signals are received over $N$ synchronized receivers, $N\geq 1$, at a sampling rate of $f_s$ (see Figure \ref{fig:pipeline}).  
Let $x$ denote the baseband transmitted signal. The set of received baseband signals $\bm{y}$ at the classifier are given by:
\begin{equation}
\bm{y}=\bm{H}x+\bm{\mathcal{N}}
\label{eq:m_c}
\end{equation}
where $\bm{y}=[y_1,\cdots,y_N]^T$ and $y_i$ represents the baseband I/Q samples at the $i$th receiver. Throughout the paper, we refer to the set of I/Q samples at given receiver as an \textit{I/Q stream}. $\bm{H}=[H_1,\cdots,H_N]^T$ where $H_i=M_ie^{j\phi_i}$, $i=1,\cdots,N$, is the baseband-equivalent complex channel between the transmitter and the $i$th receiver; $M_i$ and $\phi_i$  represent the amplitude of channel fading and its phase shift, respectively. $\bm{\mathcal{N}}=[\mathcal{N}_1,\cdots, \mathcal{N}_N]^{T}$, and $\mathcal{N}_i$ models the AWGN noise at the $i$th receiver. Over a time window $T$,  $f_s T$ I/Q samples are collected at each receiver. Then, $n$ I/Q streams are selected for further processing, where $1\leq n \leq N$ ($n$ is defined as a hyperparameter in Section \ref{sc:automl}). Next, a non-overlapping window of size $w$ is applied to these I/Q streams, and each window of I/Q samples is then min-max normalized to $[0,1]$. After normalization, inputs of size $w\times 2\times n$ are created\footnote{Samples collected over a period $T$ result in $\lfloor \frac{T f_s}{w}\rfloor$ I/Q inputs of size $w\times 2\times n$.}. We denote such an input by $I_{w\times 2\times n}$:
\begin{equation}
I_{w\times 2\times n}=
\begin{bmatrix}
	 [y^{(I)}_1[1] \cdots y^{(I)}_n[1]] & [y^{(Q)}_1[1] \cdots y^{(Q)}_n[1]]\\
	 	 [y^{(I)}_1[2] \cdots y^{(I)}_n[2]] & [y^{(Q)}_1[2] \cdots y^{(Q)}_n[2]]\\
	\vdots & \vdots \\
[y^{(I)}_1[w] \cdots y^{(I)}_n[w]]&[y^{(Q)}_1[w] \cdots y^{(Q)}_n[w]]
\end{bmatrix}	 
\label{eq:m_c}
\end{equation}
where $y^{(I)}_i[l]$ and $y^{(Q)}_i[l]$ are the $l$th in-phase and quadrature samples received from the $i$th receiver. After normalization, each $I_{w\times 2\times n}$ is fed to a CNN model to generate a class probability, and the signal type with the highest class probability is considered as the observed technology. Windows of I/Q streams are normalized to improve CNN's classification accuracy over datasets with SNR settings different than the training set's (i.e., \textit{unobseved datasets}), resulting in improved generalization performance. This is desirable since it is not feasible to gather training datasets for all possible SNR settings. Also, proper configuration of $w$ and $n$ (i.e., size of each $I_{w\times 2\times n}$) can improve classification and generalization accuracies greatly. We shed more light on this in the next section.
\section{Near-Optimal CNN Design: An AutoML Approach}\label{sc:automl}
\begin{figure}[h]
 \centering
  \includegraphics[width=\columnwidth]{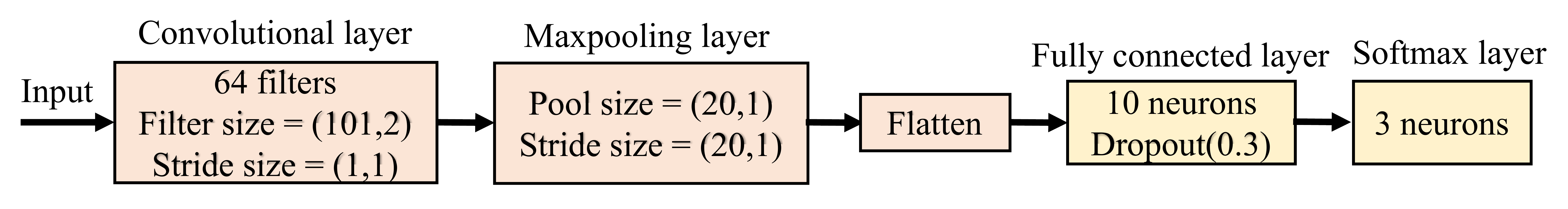}
  \caption{BCNN architecture.}
  \label{fig:basecnn}
\end{figure}

Although CNN models have been used to provide high accuracy for modulation and technology classification, significant effort is spent on fine-tuning the hyperparameters and setting various design parameters of such models. Accordingly, we seek to optimize the CNN's architectural design, preprocessing (e.g., $w$), and learning hyperparameters (e.g., learning rate). We aim at providing near-optimal values for these parameters by  forming/solving a \emph{Hyperparameter Optimization (HyperOpt)} problem. Along with \emph{Neural Architecture Search (NAS)} and \emph{Meta Learning}, HyperOpt belongs to the field of AutoML \cite{hutter2019automated}. One type of NAS, which is also applied in our work, is to treat the architectural design features, such as the number of layers, kernel sizes, etc., as hyperparameters for HyperOpt. Specifically, we explore a systematic way for HyperOpt, called \emph{Hyperband} \cite{li2017hyperband}. \\
To compare our near-optimal CNN classifiers with a manually designed CNN, we use a single-convolutional layer CNN model called \emph{Baseline CNN (BCNN)}, see Figure \ref{fig:basecnn}. We evaluate the performance of BCNN for different values of $n$ in Section \ref{sc:evaluations}. Table \ref{table:basecnn_hyper} shows BCNN's manually tuned learning hyperparameters. 

\begin{table}
\centering
\caption{BCNN's manually tuned hyperparameters.}
\resizebox{\tableSize\columnwidth}{!}{
\begin{tabular}{|c|c|}
\hline
 Activation function &  ReLU  \\ \hline
Loss  & Categorical cross-entropy  \\ \hline
Overfitting algorithm & Early stopping (patience $=6$)   \\ \hline
Batch size & 128  \\ \hline
Optimizer & Adam (learning rate $=10^{-4}$) \\ \hline
Window size ($w$) & 512 \\ \hline
Max training epochs & 100 \\ \hline

\end{tabular}}
\label{table:basecnn_hyper}
\end{table}

\subsection{Hyperparameter Optimization (HyperOpt)}\label{sc:hyperopt}
The goal of HyperOpt is to find an optimal hyperparameter configuration for a learning algorithm in a resource-efficient manner in terms of computational resources and training-time budget. To formally define a HyperOpt problem, we follow the approach in \cite{hutter2019automated} and introduce the following notation:
\begin{itemize}
\item $M$: Machine learning algorithm with $H$ hyperparameters.
\item $\Delta_h$: Domain of the $h$th hyper parameter, $h=1,\cdots,H$. The domain can consist of real values (e.g., learning rate), integers (e.g., number of fully connected layers), binary values (e.g., whether to shuffle the data at each training epoch), or a categorical value (e.g., choice of learning optimizer).
\item $\Delta:$ Hyperparameter configuration space, $\Delta=\prod_{h=1}^{H}\Delta_h$.
\item $\mu$: A vector of hyperparameter settings ($\mu\in \Delta$).
\item $M_{\mu}$: Machine learning algorithm $M$ with a hyperparameter configuration of $\mu$.
\end{itemize}

\begin{figure}[h]
 \centering
  \includegraphics[scale=0.115]{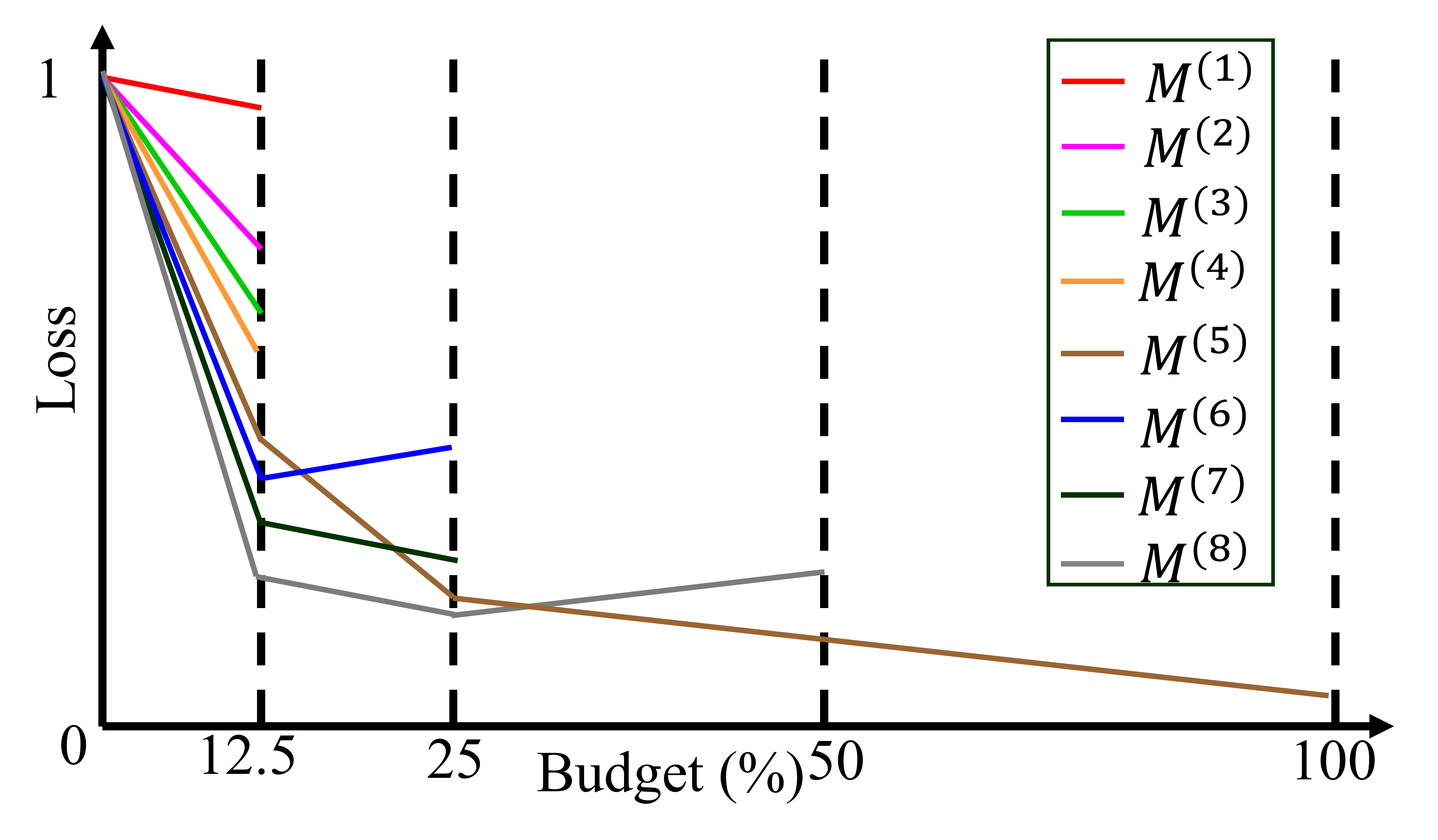}
  \caption{Successive halving for eight arbitrary hyperparameter configurations. The approach starts with eight models with different configurations and consecutively applies successive halving until only one model remains.}
  \label{fig:succesive_h}
\end{figure}

Given a dataset $D$, the goal of HyperOpt is to find an optimal hyperparameter realization $\mu^{*}$ by solving the following bi-level optimization problem:

\begin{equation}
\mu^{*}=\operatorname*{argmin}_{\mu \in \Delta} \mathbb{E}_{(D^{(\text{Train})},D^{({\text{Valid})}}) \sim D} [L(M_{\mu}, D^{(\text{Train})},D^{(\text{Valid})})].
\label{eq:hyperopt}
\end{equation} 
$L(M_{\mu}, D^{(\text{Train})},D^{(\text{Valid})})$ is the validation loss (e.g., cross-validation error) of the model generated by $M_{\mu}$, which is trained on $D^{(\text{Train})}$, and validated on $D^{(\text{Valid})}$, where $D^{(\text{Train})}$ and $D^{(\text{Valid})}$ are disjoint subsets of dataset $D$. To solve (\ref{eq:hyperopt}), we can use black-box optimization techniques (e.g., grid search, which is exponential in the number of hyperparameters), random search, population-based methods (e.g., genetic algorithms), and Bayesian optimization \cite{falkner2018bohb}. These methods are expensive to evaluate for large datasets and complex models such as CNNs. The convergence time of HyperOpt can be reduced by applying multi-fidelity optimization methods that train an algorithm with a vector of hyperparameters $\mu$ on a small subset of data, train for a few iterations, or run only on a subset of features. Multi-fidelity methods tend to minimize the low fidelity approximations of the actual loss function. Hyperband \cite{li2017hyperband} is a  pure exploration Multi-Armed Bandit (MAB)  algorithm that improves on the \emph{successive halving} algorithm, which is another bandit-based strategy for multi-fidelity optimization. Successive halving finds the best hyperparameters by initially associating computational (e.g., number of iterations to train) or storage (e.g., memory) budgets to different hyperparameter configurations of an ML algorithm/model, and trains their respective models for the number of associated training iterations. Models are then evaluated and the worst performing half of the models/algorithms are dropped, while the budget for the remaining half is doubled. The former process continues until one model/algorithm with a specific hyperparameter setting is left. See Figure \ref{fig:succesive_h} for an example of successive halving applied on eight number of hyperparameter configurations. Successive halving requires the user to decide whether to try many configurations and assign a small budget to each or to evaluate a relatively smaller set of configurations with larger assigned budgets. Assigning large budgets to poor configurations is a waste of resources, while assigning small budgets to competent configurations prematurely terminates their evaluations. Hyperband is designed to improve successive halving by systematizing budget partitioning for different hyperparameter configurations. It divides the total budget to several combinations of number of configurations and their associated budgets. The number of different configurations represent the number of arms in the MAB problem. For each arm (i.e., number of hyperparameters to consider), Hyperband calls the successive halving algorithm and keeps track of the validation loss for each evaluated hyperparameter. Finally, Hyperband recommends the hyperparameter setting with the lowest validation loss as a solution for (\ref{eq:hyperopt}). It is faster and more efficient than the successive halving algorithm, and due to its cheap low-fidelity evaluations, Hyperband has shown improved performance over random search and black-box Bayesian optimization for data subsets, feature subsets and iterative algorithms such as SGD that is used for training CNNs.

\subsection{Hyper-CNN \& Hyperparameter Settings}\label{sc:hypercnn}
\begin{figure}[h]
 \centering
  \includegraphics[scale=0.45]{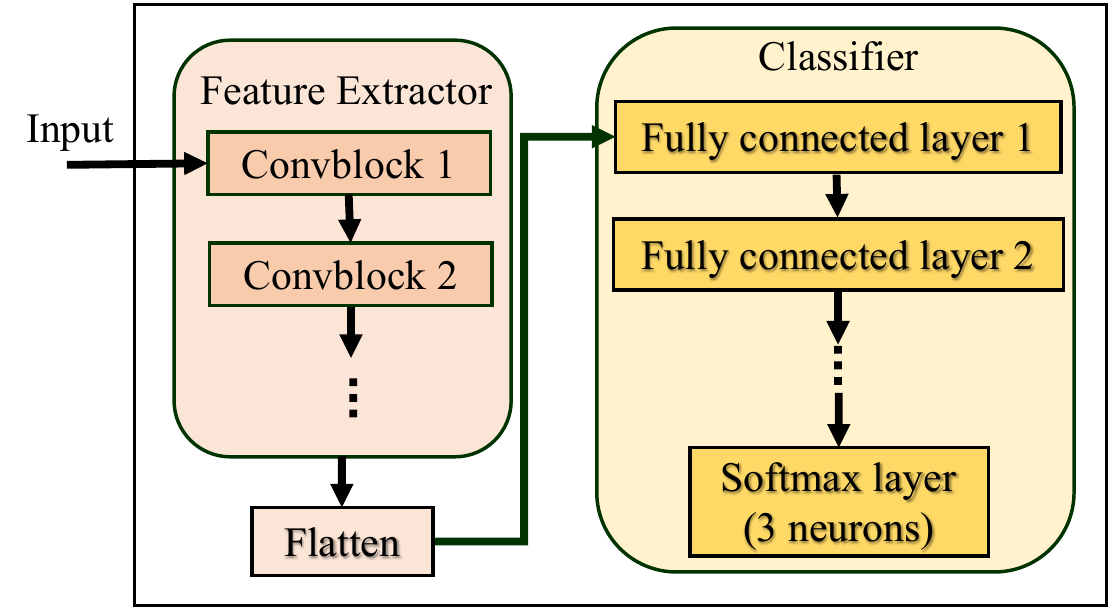}
  \caption{Hyper-CNN architecture.}
  \label{fig:hypermodel}
\end{figure}

We define the architectural design settings of our CNN model to be optimized by Hyperband as hyperparameters. We call this hypothetical architecture \textit{Hyper-CNN}. In image classification, convolutional and max-pooling layers have been used for feature extraction. These layers are often followed by a set of fully connected layers that extract nonlinear relations from the former extracted features to perform classification. We follow the same intuition and let Hyper-CNN to consist of two parts (see Figure \ref{fig:hypermodel}): \textit{Feature Extractor (FE)} and \textit{Classifier (CL)}. FE consists of one or more \textit{Convolutional Blocks (Convblocks)}. Each Convblock includes a convolutional layer, a dropout layer, and a Maxpooling layer. Within each convolutional layer, the number of filters and their sizes, the dropout rate, Maxpooling layer's pool size and stride size, and the choice of activation function are defined as hyperparameters. We also let the number of Convblocks in FE to be a hyperparameter. CL contains fully connected layers, each followed by a dropout layer with a tunable rate. We set as hyperparameters the number of fully connected layers, number of neurons in each layer, the dropout rate, and the choice of activation function. Moreover, we define as hyperparameters the choice of our training optimizer, its learning rate, batch size, and whether to shuffle training data at each epoch or not. We let the window size, $w$, and the number of selected I/Q streams, $n$, to be hyperparameters to get optimized by Hyperband, as well. Hyperband requires the domain of each hyperparameter, i.e., $\Delta_h$ to be bounded. Table \ref{table:opt_hyper} states the bounds for each hyperparameter. We fix the filter stride size of convolutional filters to $(1,1)$ and  the maximum number of training epochs to 100.

\begin{table}[h]
\centering
\caption{Hyper-CNN's hyperparameter domains (``Fully connected'' is abbreviated by FC).}
\resizebox{\tableSize\columnwidth}{!}{
\begin{tabular}{|c|c|}
\hline
 Hyperparameter  &  Domain  \\ \hline
FE: Number of Convblocks  & $\{1,2,3\}$  \\ \hline
FE: Num. filters at convolutional layer  &  $\{16,32,\cdots,160\}$  \\ \hline
FE: Filter size at convolutional layer  &  $\{16,32,\cdots,160\}\times \{2\}$  \\ \hline
FE: Pool size at the Maxpooling layer  &  $\{5,15,\cdots,75\}$  \\ \hline
FE: Stride size at the Maxpooling layer  &  $\{5,15,\cdots,60\}$  \\ \hline
CL: Num. FC layers  &  $\{1,2,3,4\}$  \\ \hline
CL: Num. neurons at FC layer  &  $\{10,20,\cdots,100\}$  \\ \hline
CL: Num. neurons at FC layer  &  $\{10,20,\cdots,100\}$  \\ \hline
FE and CL: Activation function & $\{\text{ReLU}, \text{tanh}, \text{sigmoid}\}$ \\ \hline
FE and CL: Dropout rate & $\{0,0.15,\cdots,0.6\}$ \\ \hline
Preprocessing: Window size ($w$) & $\{128,160,\cdots, 512\}$ \\ \hline
Preprocessing: Normalization & $\{\text{True},\text{False}\}$ \\ \hline
Preprocessing: Num. I/Q streams  & $\{1,2,\cdots,N\}$ \\ \hline
Learning: Optimizer & $\{\text{ADAM}, \text{SGD}, \text{RMSProp}\}$ \\ \hline
Learning: Learning rate & $\{0.01,0.001,\cdots,10^{-5}\}$ \\ \hline
Learning: Early stopping & $\{\text{True},\text{False}\}$ \\ \hline
Learning: Batch size & $\{32,64,\cdots,320\}$ \\ \hline
Learning: Shuffle data & $\{\text{True},\text{False}\}$ \\ \hline

\end{tabular}}
\label{table:opt_hyper}
\end{table}

To get near-optimal classifiers for a dataset $D$, we tune Hyper-CNN via Hyperband  using $D^{(\text{Train})}$ and $D^{(\text{Valid})}$ as described in (\ref{eq:hyperopt}). We call the tuned Hyper-CNN structure \textit{Optimal CNN (OCNN)}. Hyperband recommends different OCNNs for different datasets.

\section{Evaluation}\label{sc:evaluations}

We construct labeled datasets by generating Wi-Fi, LTE, and 5G waveforms using MATLAB WLAN \cite{MatlabWLAN}, LTE \cite{MatlabLTE}, and 5G \cite{Matlab5G} communication toolboxes.  See Table \ref{table:waveform} for a detailed description of generated waveforms, where all waveforms are $20$ MHz wide. For each waveform type, we generate 500 different waveforms by randomizing their payloads. Datasets are generated via OTA Software Defined Radio (SDR) experiments and simulations for different SNR ranges. Each generated waveform is then passed through a multi-path fading wireless channel and received by $N$ number of receivers with a sampling rate of $f_s$ MHz -- these settings vary for OTA experiments and simulations. I/Q streams are then divided into inputs of size $w\times 2\times n$ based on the settings of $w$ and $n$. Depending on its underlying technology, each $I_{w\times 2\times n}$ is labeled as a one-hot coded vector of size $1\times 3$.
\begin{table}
\centering
\caption{Waveform generation settings.}
\resizebox{\tableSize\columnwidth}{!}{
\begin{tabular}{|c|c|c|}
\hline
 Technology &   Modulation Scheme  & Code Rate \\ \hline
Wi-Fi (MC=6, 802.11 ac)  & 64-QAM  &$\sfrac{3}{4}$  \\ \hline
LTE (RC9) & 64-QAM & $\sfrac{3}{4}$   \\ \hline
5G DL FR1 (SCS= 15 kHz)  & 64-QAM & $\sfrac{1}{2}$  \\ \hline
\end{tabular}}
\label{table:waveform}
\end{table}

\begin{figure}[h]
 \centering
  \includegraphics[scale=0.07]{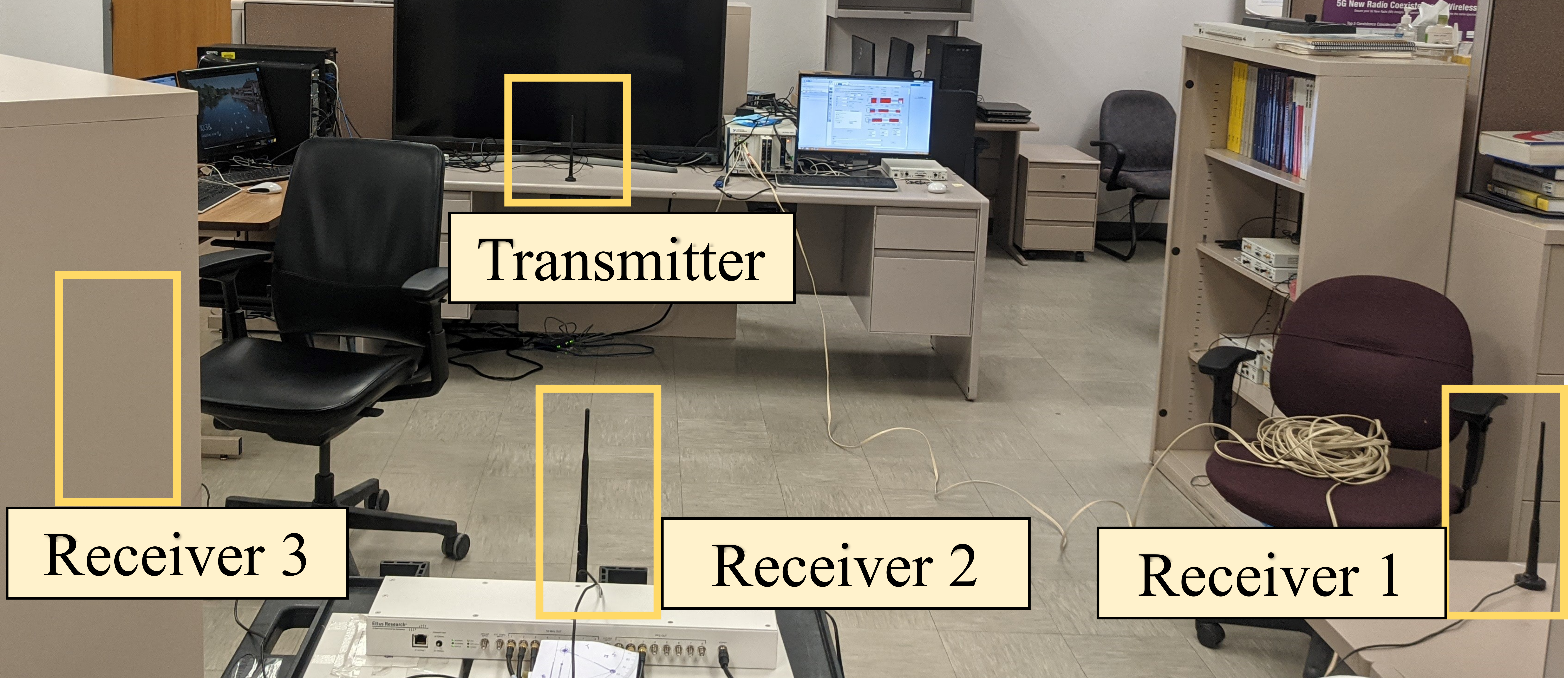}
  \caption{USRP experiment setup (Receiver 3 is inside a metallic book shelve).}
  \label{fig:usrp}
\end{figure}

Our real-world OTA experiments are based on SDRs. We use National Instruments (NI) SDRs NI-USRP 2921's. Our USRP setup consists of one transmitter and three receivers (i.e., $N=3$) each approximately 3 m away from the transmitter (see Figure \ref{fig:usrp}). Adjacent receivers are separated by 50 cm and synchronized via an Ettus Octoclock CDA-2990  with a maximum sampling rate of $f_s= $10 MHz. We observe lots of uncontrolled interference over the 5 GHz band. Hence, to not interfere with ongoing Wi-Fi transmissions, we choose a center frequency of 2.495 GHz (ISM band) for these experiments. The three receiving USRPs are connected to the same host PC via a switch. We use MATLAB SDR toolbox to transmit and receive generated waveforms. We construct multiple datasets for different transmission gains in the set $\{0, 5, 10,15,20\}$ dB which translates to average (over all three receivers) received SNR values of $\{1.5,3.5,5.5,8.5,11.5\}$ dB. Each generated dataset has 32,000 inputs for each technology.

To evaluate our approach for a larger number of receivers and a higher sampling rate, we simulate the wireless channel considering an 802.11ax™ (TGax) indoor MIMO channel model \cite{channel1}, see Table \ref{table:channel}. We generate datasets for SNR values in the set $\{-10, -5, 0, 5, 10, 15, 20\}$ dB. For each simulated dataset, we randomly select 100,000 inputs for each technology type. 

\begin{table}[h]
\centering
\caption{SIMO wireless channel settings.}
\resizebox{\tableSize\columnwidth}{!}{
\begin{tabular}{|c|c|}
\hline
 Parameter &   Value  \\ \hline
Sampling rate  & 20 Msps  \\ \hline
Delay profile & Model-D   \\ \hline
Channel bandwidth & 20 MHz  \\ \hline
Carrier frequency & 5 GHz  \\ \hline
Environmental speed & 0 m/s  \\ \hline
Transmitter-receiver distance & 3.8 m \\ \hline
Number of receive antennas ($N$) & 6 \\ \hline
Large-scale fading effects & Pathloss and shadowing \\ \hline
\end{tabular}}
\label{table:channel}
\end{table}
For each OTA and simulated dataset considered for evaluations, we use $60\%$, $20\%$, and $20\%$ of inputs for training, validation, and testing, respectively.

\subsection{Optimal CNN Architectures}

We run Hyperband after formulating the HyperOpt problem as presented in (\ref{eq:hyperopt}) for all generated datasets using their respective training and validation sets. Table \ref{table:ocnn_architectures} shows a summary of optimal hyperparameter settings for both OTA and simulated datasets. All OCNNs select ReLU as their activation function, and ADAM as their learning optimizer. They all choose to normalize their inputs, use early stopping, and shuffle the training set at the beginning of each training epoch.

\begin{table}[h]
\centering
\caption{OCCN hyperparameter settings for different datasets.}
\resizebox{\tableSize\columnwidth}{!}{
\begin{tabular}{|c|cccc|}
\hline
 SNR  & \multicolumn{4}{c|}{Hyperparameter Settings }                                                                                                                                                                              \\ \cline{2-5} 
    (dB)                          & \multicolumn{1}{c|}{$n$} & \multicolumn{1}{c|}{$w$} & \multicolumn{1}{c|}{Convblocks} & Fully Connected  \\ \hline

$1.5$ (OTA)                  & \multicolumn{1}{c|}{2}         & \multicolumn{1}{c|}{384}         & \multicolumn{1}{c|}{2}        &    3   \\ \hline
$3.5$ (OTA)                              & \multicolumn{1}{c|}{$3$}         & \multicolumn{1}{c|}{256}         & \multicolumn{1}{c|}{2}     & $2$       \\ \hline
$5.5$ (OTA)                              & \multicolumn{1}{c|}{1}         & \multicolumn{1}{c|}{448}         & \multicolumn{1}{c|}{2}         &    3    \\ \hline
$8.5$ (OTA)                               & \multicolumn{1}{c|}{3}         & \multicolumn{1}{c|}{224}         & \multicolumn{1}{c|}{1}            &    1  \\ \hline
$11.5$ (OTA)                              & \multicolumn{1}{c|}{2}         & \multicolumn{1}{c|}{448}         & \multicolumn{1}{c|}{1}          &     2  \\ \hline

$-10$ (Simulations)                   & \multicolumn{1}{c|}{6}         & \multicolumn{1}{c|}{320}         & \multicolumn{1}{c|}{1}        &    1    \\ \hline
$-5$ (Simulations)                   & \multicolumn{1}{c|}{1}         & \multicolumn{1}{c|}{416}         & \multicolumn{1}{c|}{1}        &    2    \\ \hline
$0$ (Simulations)                  & \multicolumn{1}{c|}{5}         & \multicolumn{1}{c|}{448}         & \multicolumn{1}{c|}{2}        &    1    \\ \hline
$5$ (Simulations) & \multicolumn{1}{c|}{$5$} & \multicolumn{1}{c|}{228}         & \multicolumn{1}{c|}{2}     & $1$   \\ \hline
$10$ (Simulations) & \multicolumn{1}{c|}{3} & \multicolumn{1}{c|}{352}         & \multicolumn{1}{c|}{1}         &    2    \\ \hline
$15$ (Simulations) & \multicolumn{1}{c|}{6}  & \multicolumn{1}{c|}{320}         & \multicolumn{1}{c|}{2}            &    3   \\ \hline
$20$ (Simulations) & \multicolumn{1}{c|}{6}   & \multicolumn{1}{c|}{320}         & \multicolumn{1}{c|}{2}          &     3  \\ \hline

\end{tabular}}
\label{table:ocnn_architectures}
\end{table}

\begin{figure}[h]
 \centering
  \includegraphics[scale=0.13]{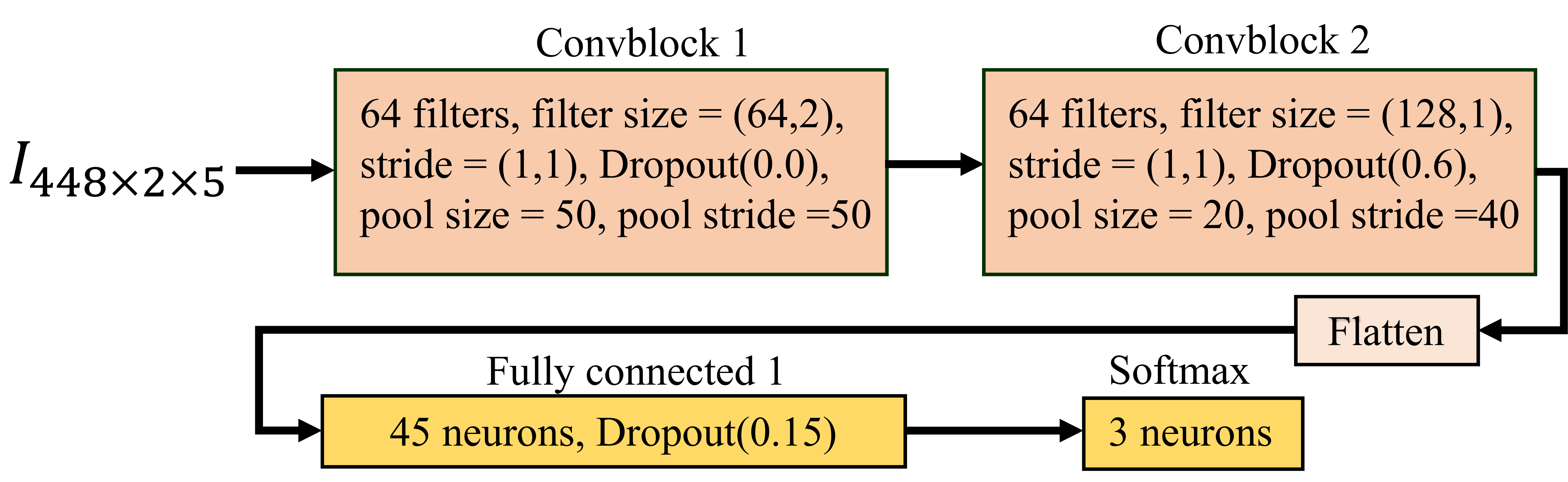}
  \caption{OCNN architecture for simulation dataset with an SNR value of 0 dB.}
  \label{fig:OCNN0}
\end{figure}

It can be concluded from the table that most OCNN structures prefer having multi-channel inputs, i.e., selecting I/Q streams from multiple receivers, rather than using single-channel inputs. Moreover, on average, OCNNs choose $w$ to be $347$, select 1.58 number of Convblocks and 1.83 number of fully connected layers. Due to space limit we can not show other detailed architectural settings (e.g., filter size). However, Figure \ref{fig:OCNN0} shows the detailed architecture of OCNN tuned for the simulation dataset with an SNR of 0 dB.  In Section \ref{sc:gen_ac}, we show that this OCNN architecture achieves the highest average generalization performance compared to other OCNNs.

\subsection{OCNN vs. BCNN }
\begin{figure}[h]
\centering
  \subfigure[]{\begin{tikzpicture}[scale=0.4, transform shape]
\begin{axis}[title={},xlabel={SNR (dB)},
    ylabel={Classification Accuracy ($\%$)},
    xmin=1.5, xmax=11.5,
    ymin=30, ymax=100,
    xtick={1.5,3.5,5.5,8.5,11.5},
    ytick={30,40,60,80,100},
    yticklabels ={30,40,60,80,100},
    legend style={fill=none,at={(axis cs:1.5,145)},anchor=north west}
    ,nodes={scale=2, transform shape}]

\addplot[color=black,mark=*]coordinates{(1.5,38.1)(3.5,45.83)(5.5,60.06)(8.5,61.22)(11.5,47.84)};
\addlegendentry{BCNN ($n=1$)}

  \addplot [color=magenta, mark=star] coordinates {(1.5,56.82)(3.5,40.17)(5.5,56.91)(8.5,55.84)(11.5,68.57)};
 \addlegendentry{BCNN ($n=2$)};
  \addplot [color=olive, dashed,  line width=2pt] coordinates {(1.5,54.42)(3.5,45)(5.5,60.2)(8.5,57.41)(11.5,65.65)};
 \addlegendentry{BCNN ($n=3$)};
   \addplot [color=blue, mark=square] coordinates {(1.5,61.65)(3.5,70.78)(5.5,70.74)(8.5,73.76)(11.5,83.59)};
 \addlegendentry{OCNN};
\end{axis}
\end{tikzpicture}\label{OCNN_vs_BCNN_ota}}\subfigure[]{\begin{tikzpicture}[scale=0.4, transform shape]
\begin{axis}[title={},xlabel={SNR (dB)},
    xmin=-10, xmax=20,
    ymin=30, ymax=100,
    xtick={-10,-5,0,5,10,15,20},
    xtick={-10,-5,0,5,10,15,20},
    ytick={30,40,60,80,100},
    yticklabels ={30,40,60,80,100},
    legend style={fill=none,at={(axis cs:-10,145)},anchor=north west}
    ,nodes={scale=2, transform shape}]

\addplot[color=black,mark=*]coordinates{(-10,33.2)(-5,33.2)(0,48.99)(5,71.36)(10,89.32)(15,95.74)(20,99.5)};
\addlegendentry{BCNN ($n=1$)}
 
  \addplot [color=magenta, mark=star] coordinates {(-10,33.18)(-5,33.19)(0,33.23)(5,43.19)(10,60.82)(15,81.87)(20,72.22)};
 \addlegendentry{BCNN ($n=3$)};

  \addplot [color=olive, dashed,  line width=2pt] coordinates {(-10,33.19)(-5,33.23)(0,33.18)(5,33.25)(10,36.42)(15,42.92)(20,55.89)};
 \addlegendentry{BCNN ($n=6$)};
 
   \addplot [color=blue, mark=square] coordinates {(-10,55.68)(-5,66.52)(0,97.68)(5,99.22)(10,99.82)(15,99.93)(20,99.95)};
 \addlegendentry{OCNN};
\end{axis}
\end{tikzpicture}\label{OCNN_vs_BCNN_sim}}

  \caption{ BCNN and OCNN classification accuracy vs. SNR for (a) OTA USRP and (b) simulation datasets.}
    \label{fig:BCNN_usrp}

  \end{figure}
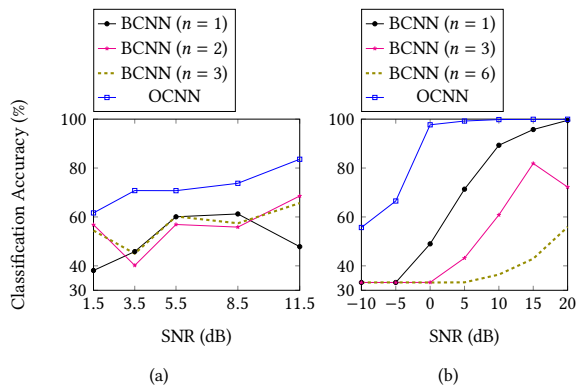

 In Figures \ref{OCNN_vs_BCNN_ota} and \ref{OCNN_vs_BCNN_sim}, we show classification accuracy of BCNN with different number of  selected I/Q streams and OCNN architectures vs. received SNR, for OTA experiments and simulations, respectively. On average, OCNNs improve classification accuracy of best-performing BCNNs by  $24.58\%$ and $46.15\%$, for OTA experiments and simulations, respectively. Also it can be concluded that increasing the number of selected I/Q streams does not necessarily increase the accuracy and it is better to define it as a hyperparameter to be optimized by Hyperband. Moreover, OCNNs' classification accuracy over simulation datasets shows that the classification task becomes easier to solve as SNR is increased; whereas, over low SNR values, OCNNs improves accuracy by a greater margin.

\subsection{Generalization over Unobserved Datasets}\label{sc:gen_ac}
\begin{figure}
 \centering
\begin{tikzpicture}[scale=0.8, transform shape]
    \pgfplotsset{ybar stacked, ymin=0, ymax=100, xmin=0.5, xmax=7.5,  xticklabels={,,$\langle2\,2\rangle$,,$\langle4\,2\rangle$,,$\langle15\,2\rangle$,,$\langle15\,2\rangle$ ,, $\langle15\,2\rangle$,, $\langle15\,2\rangle$,, $\langle15\,2\rangle$}
    ,
    xlabel={$S_w$},
    ylabel={Generalization Accuracy ($\%$)}
    }

 \begin{axis}[bar shift=-8pt,
    bar width=6pt,
legend style={at={(axis cs:0.5,120)},anchor=north west},
         xlabel={Training set's SNR (dB)},
         ylabel={Generalization Accuracy ($\%$)},
         xticklabels={,,$-10$,$-5$,$0$,$5$,$10$,$15$,$20$}]
    \addplot [fill=white] table [x index = 0, y index = 1] {allSnrNotNormalize.dat};

  \small  \legend{All SNRs (not normalized)}
    
  \end{axis}
  
 \begin{axis}[bar shift = 0pt,
     bar width=6pt,
      xlabel={},
          ylabel={},
          xticklabels={,,,,,,,,,},
             yticklabels={},
   legend style={at={(axis cs:0.5,110)},anchor=north west}]
    \addplot [fill=yellow,postaction={pattern=north east lines}] table [x index = 0, y index = 1] {allSnrNormalize.dat};
   \small   \legend{All SNRs (normalized)}
  \end{axis}
  
 \begin{axis}[bar shift = 8pt,
     bar width=6pt,
      xlabel={},
       ylabel={},
                 xticklabels={,,,,,,,,,},
                    yticklabels={},
   legend style={at={(axis cs:3.8,110)},anchor=north west}]
          \addplot [fill=red, postaction={pattern=dots}] table [x index = 0, y index = 1] {adjacentSnrNormalize.dat};

   \small \legend{Adjacent SNRs (normalized)}
  \end{axis}
    \end{tikzpicture}  
      \caption{Average generalization accuracy with/without normalized inputs vs. SNR setting which the model was trained on.}
          \label{fig:sim_OCNN_unobs}
\end{figure}
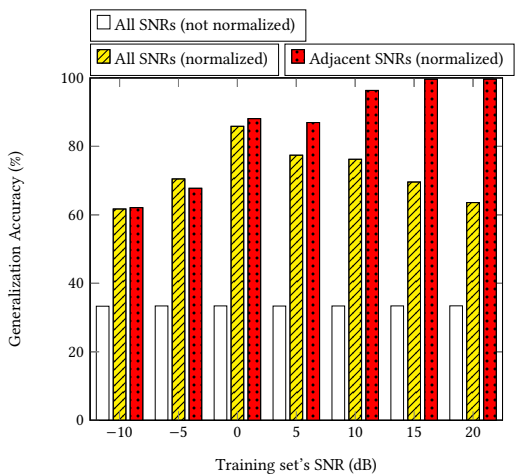

Generalization accuracy is the classification accuracy of a model when it is tested on datasets with SNR values other than the training set's SNR. In Figure \ref{fig:sim_OCNN_unobs} we show the average generalization accuracy vs. the SNR of the training using simulations. It can be seen that normalization improves the generalization accuracy by $108.05\%$, on average. We also show the generalization accuracy evaluated on datasets with SNR values that are 5 dB higher or lower than the training set's (except for $-10$ and $20$ dB, where we only test on $-5$ and $15$ dB, respectively). It can be seen that on average models are  $19.7 \%$ more accurate for close SNR values.

\section{Conclusions}\label{sc:conclusion}
In this work, we considered the protocol classification problem of raw received I/Q samples for Wi-Fi, LTE-LAA, and 5G NR-U. We presented a  multi-receiver technology classification method that uses a CNN model to classify multiple streams of I/Q samples collected from differently located synchronized receivers. We provided an AutoML approach to jointly optimize CNN architectures, preprocessing hyperparameters, such as number of selected I/Q streams, and learning configurations (e.g., choice of optimizer). Our USRP  experiments showed $20.58\%$ gain when optimal CNNs are used instead of manually tuned CNNs. We also showed that increasing the number of I/Q streams does not necessarily increase classification accuracy and it should be optimized as a part of our HyperOpt problem. For our simulated datasets, we found that normalizing I/Q samples improves generalization accuracy by $108.05\%$.

\section{Acknowledgments}
This research was supported by the U.S. Army Small Business Innovation Research Program Office and the Army Research Office under Contract No. W911NF-21-C-0016, by NSF (grants CNS-1563655, CNS-1731164, and IIP-1822071), and by the Broadband Wireless Access \& Applications Center (BWAC). Any opinions, findings, conclusions, or recommendations expressed in this paper are those of the author(s) and do not necessarily reflect the views of NSF or ARO.

\bibliographystyle{ieeetr}
\bibliography{The_bilbiography}

\end{document}